\documentclass[journal]{IEEEtran}
\usepackage{multicol}
\usepackage{subfig}
\usepackage{stfloats}
\usepackage{hyperref}
\usepackage{url}

\usepackage {graphicx}
\usepackage{array}
\usepackage{amssymb}
\usepackage{dsfont}

\ifCLASSINFOpdf
\else
\fi
\begin{document}
%
\title{Feature Fusion using Extended
Jaccard Graph and Stochastic Gradient Descent for Robot}
%
%
%

\author{Shenglan~Liu, Muxin~Sun, Wei~Wang, Feilong~Wang
\thanks{
Shenglan Liu  and Wei Wang are with Faculty of Electronic Information and Electrical Engineering, Dalian University of Technology, Dalian, Liaoning, 116024 China. Feilong Wang is with the School of Innovation and Entrepreneurship, Dalian University of Technology, Dalian, Liaoning, 116024 China. Muxin Sun is with the State Key Laboratory of Software Architecture (Neusoft Corporation).  e-mail: ( liusl@dlut.edu.cn).}
}

%
%

\markboth{Journal of \LaTeX\ Class Files}%
{Shell \MakeLowercase{\textit{et al.}}: Bare Demo of IEEEtran.cls for IEEE Journals}
%



\maketitle

\begin{abstract}
Robot vision is a fundamental device for human-robot interaction and robot complex tasks. In this paper, we use Kinect and propose a feature graph fusion (FGF) for robot recognition. Our feature fusion utilizes RGB and depth information to construct fused feature from Kinect. FGF involves multi-Jaccard similarity to compute a  robust graph and utilize word embedding method to enhance the recognition results. We also collect DUT RGB-D face dataset and a benchmark datset to evaluate the effectiveness and efficiency of our method. The experimental results illustrate FGF is robust and effective to face and object datasets in robot applications.
\end{abstract}


%
\IEEEpeerreviewmaketitle

\section{Introduction}
The object recognition is one of the important problems in machine vision
and essential capabilities for social robot in real word environments.
Object recognition in real word is a challenged problem because of
environment noisy, complex viewpoint, illumination change and shadows. 2D
camera always cannot deal with such hard task. Kinect \cite{13} released
promising a new approach to help compliant~hand designing \cite{1}, recognize
objects and human (emotions) for robot. The characteristic of the new Kinect
2.0 release are list as follows (see figure 1):

RGB Camera£ºTake the color image/video in the scope of view.

IR Emitters: When actively projected Near Infrared Spectrum (NIS) irradiates
to rough object or through a frosted glass, spectrum will distort and form
random spots (called speckle) that can be read by an infrared camera.

\begin{center}
\begin{figure}[htbp]
\centerline{\includegraphics[width=3.5 in]{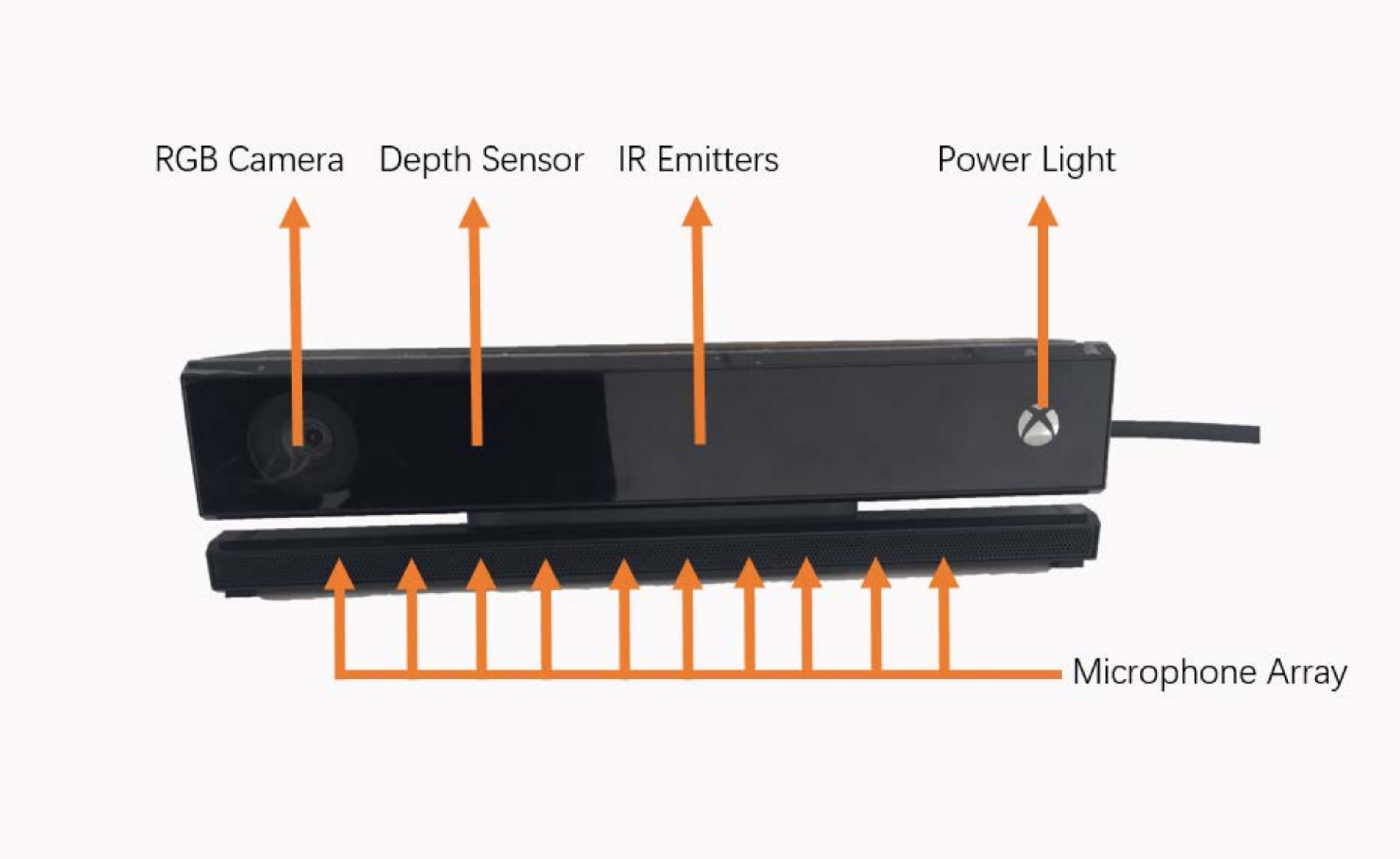}}
\label{fig1}
\caption{ The structure of Kinect V2}
\end{figure}
\end{center}

Depth Camera£ºAnalyze infrared spectrum, and create RGB-Depth (RGB-D) images
of human body and objects in the visual field.

Microphone Array£ºEquip built-in components (e.g. Digital Signal Processor
(DSP)) to collect voice and filter background noise simultaneously. The
equipment can locate the sound source direction.

Kinect (RGB-D sensor) can capture a color image and corresponding depth
information of each pixel for real word objects and scenes synchronously.
The color and depth images are complementary information for real-word
tasks. A significant number of applications are exploited by RGB-D sensor.
It can be referred to object detection \cite{2, 10, 12}, human motion analysis
\cite{3, 11}, object tracking \cite{9}, and object/human recognition [4-8, 14, 15, 33]
etc.. In this paper, we only discuss recognition problem by using RGB-D
sensor. The recent feature-based object recognition methods mainly fall into
three categories: converting 3D point clouds, jointing 2D and depth image
features and designing RGB-D feature.

To extract 3D object feature, Bo et al.\cite{16}  utilizes depth information and
maps to generate 3D point clouds. Motivated by local feature in RGB image,
depth kernel feature can be extracted for 3D point clouds. However, feature
of 3D point clouds will suffer from noisy and the limited views while only
one view/similar views is available and noisy involved. Jointing 2D and
depth image features is a flexible approach for RGB-D vision learning. This
relies on many excellent 2D image descriptors which are proposed in computer
vision.

Local Binary Patterns (LBP) \cite{17,18} and Histogram of Gradient (HOG) \cite{19}, which according to the texture and edge of image respectively. Recently, motivated by the visual perception mechanisms for image retrieval, perceptual uniform descriptor (PUD)\cite{20} achieves high performance by involving human perception. PUD defined the perceptual-structures by the similarity of edge orientation and the colors, and introduced structure element correlation statistics to capture the spatial correlation among them. On the contrary, local image descriptors focus on describing local information which includes edge and gradient information etc.. Lowe et al. \cite{21} introduced a classical local descriptor called scale-invariant feature transform (SIFT), which aims to detect and describe local neighborhoods closing to key points in scale space. SIFT and HOG are both can be included in Bag of Words (BOW) framework as image descriptor. Gabor wavelets \cite{22} have been also applied to image understand, which due to vision similarity of human and Gabor wavelets. HMAX-based image descriptor \cite{23} according to the hierarchical visual processing in the primary visual cortex (V1) can get promising results in vision tasks. The above image descriptors can be selected to process the RGB and depth image respectively. However, most jointing features are incompatible, which means difficult to choose a suitable weight for jointing.

To full use of the 2D image descriptor, we intend to utilize fusion approach
to get 3D object feature for further RGB-D vision learning. Few feature
fusion methods are proposed in RGB-D recognition to our best knowledge. The
most popular approach is multi-view graph learning \cite{28}. This approach can get
good performance while only classifies a small quantity of objects.

\begin{center}
\begin{figure}[htbp]
\centerline{\includegraphics[width=3.50in]{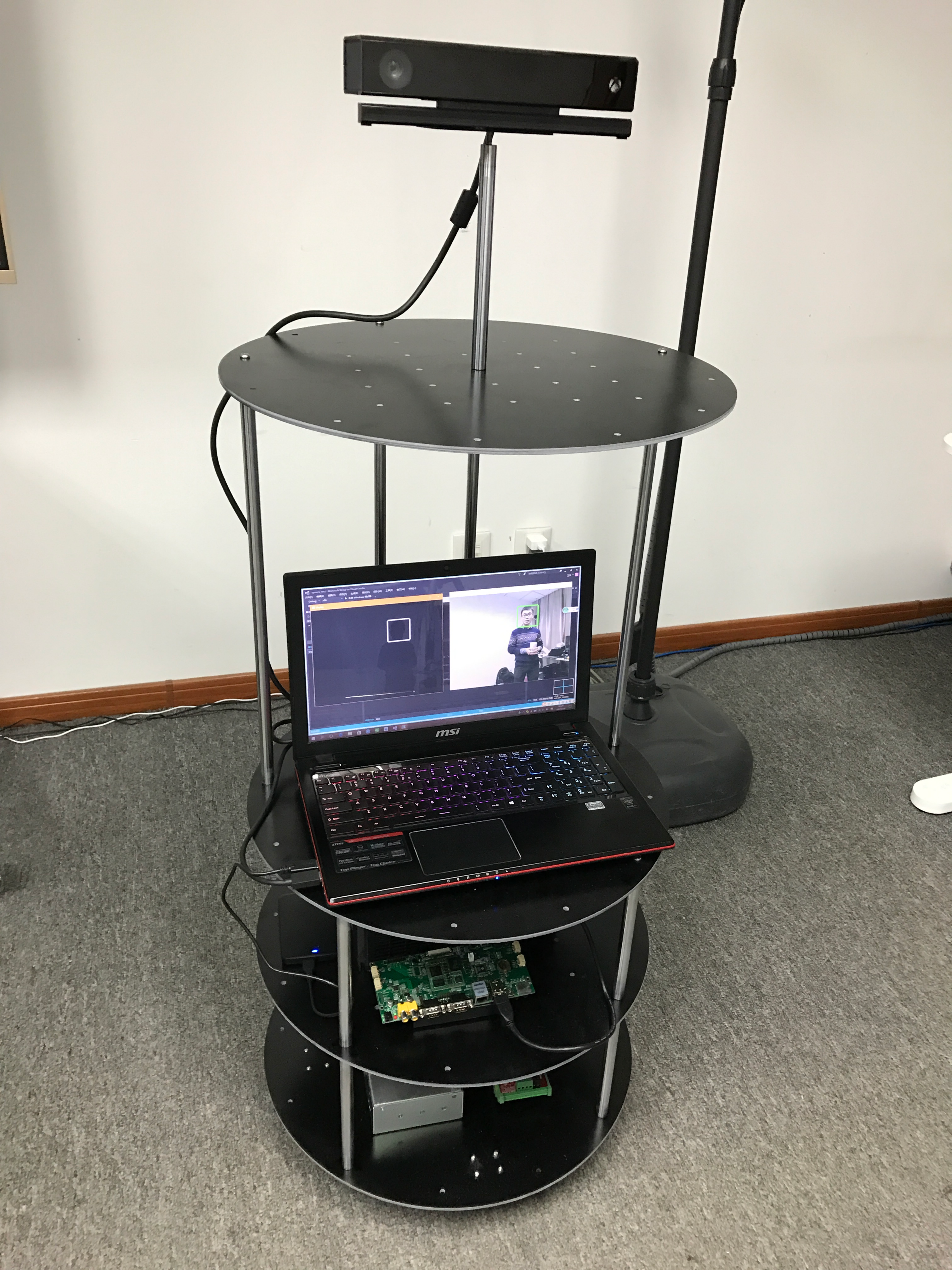}}
\label{fig2}
\caption{A simple Kinect-based robot}
\end{figure}
\end{center}

In this paper, a new feature graph fusion (FGF) method is proposed for RGB
and depth images. We first utilize Jaccard similarity to construct a graph
of RGB and depth images, which indicates the similarity of pair-wise images.
Then, fusion feature of RGB and depth images can be computed by our Extended
Jaccard Graph (EJG) using word embedding method. Our
Feature Graph Fusion can get better performance and efficiency in RGB-D
sensor for robots. Our simple Kinect-based robot is show in figure 2.

\section{BOW based on SIFT and CNN-RNN}

In this section, we will introduce the fundamental features which are used
in our paper. As describe in introduction section, local and Bio-Inspired
features always can achieve better results than other type ones. Subsection
2.1 and 2.2 introduce BOW and CNN-RNN features respectively.

\subsection{BOW based on SIFT}

Scale Invariant Feature Transform (SIFT) was first introduced to extract
distinctive local features for image matching. This feature has great
discriminative performance and robustness and has been widely applied for
various vision tasks. It is significantly invariant to translation, rotation
and rescaling of images, and also has certain robustness to change in 3D
viewpoint and illumination. There are two major stages in SIFT for
maintaining the superior properties: detector and descriptor.

SIFT detector aims at find out the key points or regions in the Gaussian
scale space. Since natural images from camera or other devices tend to be
sampled from different views, it is necessary to construct scale space
pyramid to simulate all the possible scales for identifying accurately the
locations and scales of key points. And then the locations can be determined
using local extreme detection in the difference-of-Gaussian scale space.
Some low contrast or edge responses need to be further removed due to their
less discrimination.

SIFT descriptor is computed using the image gradients in the neighborhood of
key point. In order to maintain rotation invariance, the descriptors need to
be rotated relative to the key point orientation. And then by computing the
gradient information in the neighborhood of key point, the descriptor
characterizes the orientation distribution around the key point which is
distinctive and partially robust to illumination or 3D viewpoint. So for
each key points detected above, a 128-D feature vector is created to extract
the local discriminative information.

Even though SIFT has superior performance in local feature description and
matching without high efficiency, it is still not appropriate to be used for
analyzing the holistic image feature directly considering the large amount
of key points in each image. So by introducing Bag-of-Word model to be
combined with SIFT, it takes the main SIFT vectors as the basic words, which
preserves the great distinction of SIFT. The main procedure for this
strategy is to find out the cluster centers as words by pre-training with
k-means, and then create the words vectors by assigning the whole SIFT key
points into the nearest word. Normally, these cluster centers contains the
discriminative patches among images. So the selected number of words also
plays an important part in image representation. The large words might
create elaborate features which describe more detailed information, while
the small words mainly consider coarse distribution of these SIFT
descriptors.

\begin{figure*}[htbp]
\centerline{\includegraphics[width=5.5 in]{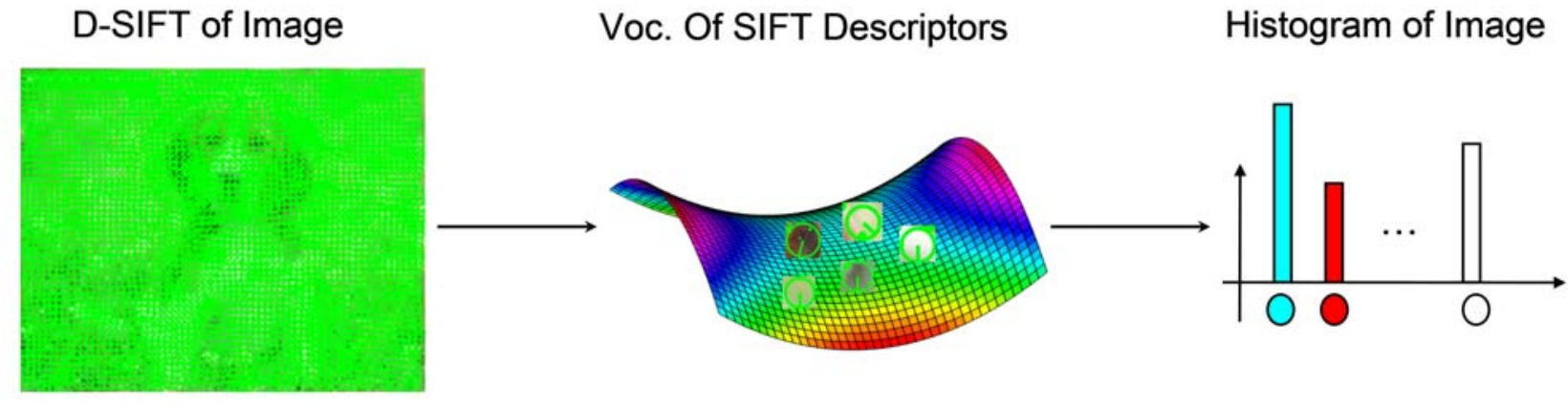}}
\label{fig3}
\caption{Process of BOW using D-SIFT}
\end{figure*}

In D-SIFT, SIFT descriptor is created in the whole image region without
detecting the key points in scale-space. Instead of smoothing the images by
Gaussian kernel in scale-space, the image is pre-smoothed before feature
description. So it is much faster than standard SIFT since the key point
detection tends to be very time-consuming in large-scale image
understanding. The main process of BOW using D-SIFT is concluded as Fig.
3.

\subsection{CNN-RNN (CRNN)}

Richard Socher et. al \cite{r1}  proposed CNN-RNN model which has 2 steps: 1) learning the CNN filters in an unsupervised way by clustering random patches and then feeding these patches into a CNN layer. 2) using the resulting low-level, translationally invariant features generated from the first step to compose higher order features that can then be used to classify the images with RNNs.

First, random patches are extracted into two sets: RGB and depth. Then, each set of patches is normalized and whitened. K-means classifier is used to cluster patches for pre-processing.

Second, a CNN architecture is chosen for its translational invariance properties to generate features for the RNN layer. The main idea of CNNs is to convolve filters over the input image. The single layer CNN is similar to the one proposed by Jarrett et. al \cite{r17} and consists of a convolution, followed by rectification and local contrast normalization (LCN) \cite{r17,r18,r14}. Each image of size (height and width) $d_I$ is convolved with $K$ square filters of size $d_P$ , resulting in $K$ filter responses, and each of which is with dimensionality $d_I-d_P+1$. After that, the image averagely pools them with square regions of size $d_\ell$ and a stride size of s, to obtain a pooled response with width and height equal to $r = (d_I-d_\ell)/s+1$. So the output X of the CNN layer applied to one image is a $K\times r\times r$ dimensional 3D matrix. The same procedure is applied to both color and depth images separately.

The idea of recursive neural networks \cite{r19, r9} is to learn hierarchical feature representations by applying the same neural network recursively in a tree structure. In the case of CNN-RNN model, the leaf nodes of the tree are K-dimensional vectors (the result of the CNN pooling over an image patch repeated for all $K$ filters) and there are $r^2$ of them.

 It starts with a 3D matrix $X\in \mathbb{R}^{K\times r\times r}$ for each image (the columns are K-dimensional), and defines a block to be a list of adjacent column vectors which are merged into a parent vector $p\in R^K$. For convenience, only square blocks with size $K\times b\times b$ are employed. For instance, if vectors are merged in a block with $b = 3$, it will output a total size $128\times 3\times 3$ and a resulting list of vectors $(x_1,\cdots, x_9)$. In general, their are $b^2$ vectors in each block. The neural network where the parameter matrix $W\in \mathbb{R}^{2K}$, $f$ is a nonlinearity such as $\tanh$. Generally, there will be $(r/b)^2$ parent vectors $p$, forming a new matrix $P_1$. The vectors in $P_1$ will again be merged in blocks just as those in matrix $X$ with the same tied weights resulting in matrix $P_2$.

\section{Feature Graph Fusion}

In this section, EJG will propose by Jaccard similarity for robust graph
construction which is important to our FGF in this paper.

\subsection{Extended Jaccard Graph}

We use extended Jaccard graph to construct a fused graph to compute feature fusion. The detail of graph fusion are described in this subsection.

We first define $H \in \mathbb{R}^{D \times n}$ as the original image set. Let $h_q $ denote the query,
and $N_k \left( {h_q } \right) = \left\{ {h_{qc} } \right\}$ represent the
KNNS{\footnote{KNNS indicates the $k$ nearest neighborhood of a sample.}} of $h_q $, $c = 1, \cdots ,k$. $N_k \left( {h_q } \right)$ is the
original ranking list which returns top-$k$ images of $x^q$. Similar to $h_q
$, we denote $N_{k_1 } \left( {h_{qc} } \right) = \left\{ {h_{qc}^i }
\right\}$ as the KNNS of $h_{qc} $, $i = 1, \cdots ,k_1 $, $N_{k_2 } \left(
{h_{qc}^i } \right)$ as the KNNS of $h_{qc}^i $. Jaccard coefficient
($J\left( { \cdot , \cdot } \right))$ is set to measure the similarity
of $h_{qc} $ and $h_q $ as follows:

\begin{equation}
\label{eq1}
J\left( {h_{qc} ,h_q } \right) = \frac{\left| {N_{k_1 } \left( {h_{qc} }
\right) \cap N_k \left( {h_q } \right)} \right|}{\left| {N_{k_1 } \left(
{h_{qc} } \right) \cup N_k \left( {h_q } \right)} \right|}
\end{equation}

The information in $J\left( {h_{qc} ,h_q } \right)$ is more than that in
norm measure of $h_{qc} $ and $h_q $. In construction of the graph, the edge weight
between $h_{qc} $ and $h_q $ is denoted by $w\left( {h_{qc} ,h_q } \right)$
in Eq.(\ref{eq1}), where $\alpha $ is a decay coefficient.

To avoid outliers in $N_{k_1 } \left( {h_{qc} } \right)$, we consider
comparing $N_{k_1 } \left( {h_{qc} } \right)$ with $N_{k_2 } \left(
{h_{qc}^i } \right)$ by the similar process of $h_{qc} $ and $h_q $ as
follows:

\begin{equation}
\label{eq2}
J\left( {h_{qc} ,h_{qc}^i } \right) = \frac{\left| {N_{k_1 } \left( {h_{qc}
} \right) \cap N_{k_2 } \left( {h_{qc}^i } \right)} \right|}{\left| {N_{k_1
} \left( {h_{qc} } \right) \cup N_{k_2 } \left( {h_{qc}^i } \right)}
\right|}
\end{equation}

We utilize the results of Eq. (\ref{eq2}) to define  in Eq.(\ref{eq3}). If the value of
${w}'\left( {h_{qc} ,h_{qc}^i } \right)$ is small enough,
$h_{qc} $ is the outlier to query.

\begin{equation}
\label{eq3}
{w}'\left( {h_{qc} ,h_{qc}^i } \right) = \left\{ {{\begin{array}{*{20}c}
 1 \hfill & {\left( {J\left( {h_{qc} ,h_{qc}^i } \right) > 0} \right) \wedge
\left( {h_{qc}^i \in N_{k_1 } \left( {h_{qc} } \right)} \right)} \hfill \\
 0 \hfill & {else} \hfill \\
\end{array} }} \right.
\end{equation}

Then, the weight of $h_{qc} $ and $h_q $ can be computed by Eq. (\ref{eq4}) as
follows£º

\begin{equation}
\label{eq4}
w\left( {h_{qc} ,h_q } \right) = \sum\limits_{h_{qc}^i \in N_{k_1 } \left(
{h_{qc} } \right)} {{w}'\left( {h_{qc} ,h_{qc}^i } \right)}
\end{equation}

In order to obtain the complementary information of RGB and depth image features to improve the accuracy of machine/robot recognition, we need to fuse multi-feature of images. We denote $V$ as node, $E$ as edge and $w$ as weight in image graph. Assuming RGB and depth features have been extracted from an object. Then RGB and depth graphs can be constructed by Extended Jaccard Graph in reference \cite{24}. In graph fusion methods, the RGB feature graph defines as $G^{rgb} = \left( {V^{rgb}, E^{rgb}, w^{rgb}} \right)$, and depth feature graph can be denoted by $G^{depth} = \left( {V^{depth}, E^{depth}, w^{depth}} \right)$. Multi-feature graph can be expressed by $G = \left({V, E, w} \right)$ which satisfies three constrains as follows: 1) $V =V^{rgb} \bigcup V^{depth} $; 2) $E = E^{rgb} \bigcup E^{depth}$; 3) $w\left( {\hat {x},x^q} \right) = {w^{rgb}\left({\hat {x},x^q} \right)} \bigcup {w^{depth}\left({\hat {x},x^q} \right)}$. The fusion graph $G$ can be treat as the relationships between images in dataset. We can also get the final fusion feature on $G$ in the next subsection.

\subsection{Fusion Feature by Word Embedding}

We fuse RGB weight affinity matrix $w^{rgb}$ and depth weight affinity
matrix $w^{depth}$ as affinity matrix $W = \left[ {W_{ij} } \right]$, where
$i,j \in V$ are denotes in subsection \uppercase\expandafter{\romannumeral3}A. Then, we can get the normalized
neighborhood affinity matrix$W = \left[ {W_{ij} } \right]$, where $i,j \in
V$. $W_{ij} $ can be expressed using a Gaussian kernel as follows.

\begin{equation}
\label{eq5}
W_{ij} = \left\{{{
\begin{array}{*{20}c}
{\frac{\exp \left( { - \frac{W_{ij}}{2\sigma_{i} ^2}}
\right)}{\sum\nolimits_{{j}' \in N_k \left( i \right)} {\exp \left( { -
\frac{W_{i{j}'}}{2\sigma _{i} ^2}} \right)} },} \hfill & {j \in N_k \left( i
\right)} \hfill \\
 {0,} \hfill & {else} \hfill \\
\end{array} }}\right.
\end{equation}

\noindent
where $\sigma _i $ is the bandwidth parameter of Gaussian kernel, we denote
$\sigma _i $ by variance of the i-th row.

The fused features are implicit expression in the normalized neighborhood
affinity matrix $W = \left[ {W_{ij} } \right]$. We use the following
optimization models to get the fused features.

\begin{equation}
\label{eq6}
\mathop {\max }\limits_F \prod\nolimits_{i,j \in V} {\left( {\frac{\exp \left(
{f_i^T f_j } \right)}{\sum\nolimits_{j \in V} {\exp \left( {f_i^T f_j }
\right)} }} \right)} ^{W_{ij} }
\end{equation}

, where $\{ f_1 ,f_2 , \cdots , f_n \} \in {\mathbb{R}}^{d\times
n} $, $f_i$ is the fused feature of the $i$-th RGB-D image pair. We change the likelihood function into log function as follows.

\begin{figure*}
\begin{equation}
\label{eq7}
\ell = \sum\nolimits_{i,j \in V} {W_{ij} \cdot \log \left( {\frac{\exp
\left( {f_i^T f_j } \right)}{\sum\nolimits_{j \in V} {\exp \left( {f_i^T f_j
} \right)} }} \right)} \approx \frac{\sum\nolimits_{i \in V}
{\sum\nolimits_{j \in s\left( i \right)} {\log \left( {\frac{\exp \left(
{f_i^T f_j } \right)}{\sum\nolimits_{j \in V} {\exp \left( {f_i^T f_j }
\right)} }} \right)}}}{M}
\end{equation}
\end{figure*}

, where $s(i)$ indicates sampling $M$ times according to the distribution
function which generates by the $i$-th row of $W$. The optimization function
can get $F$ by using word embedding model\cite{25,26}.

\section{Experimental Results and Analysis}

In this section, we use two datasets to evaluate our feature fusion method.
We first introduce the parameters and details of the two dataset. Then, the
results and its analysis of the experiments are list in subsection 4.2.

\subsection{Details of Datasets}

The dataset 1 and dataset 2 are collected by Kinect V1 and V2 respectively.
The difference between V1 and V2 are listed in Table 1. Dataset 1 is
recorded by Kinect V2 and Dataset 2 is given by Kinect V1. The two datasets
are described as follows.

\begin{table}[htbp]
\centering
\caption{Comparison of Kinect V1 and V2}
\begin{tabular}
{|c|c|c|c|}
\hline
\multicolumn{2}{|c|}{Parameter } &
Kinect V1 &
Kinect V2  \\
\hline
\raisebox{-1.50ex}[0cm][0cm]{Color }&
Resolution&
640$\times $480&
1920$\times $1080 \\
\cline{2-4}
 &
fps&
30fps&
30fps \\
\hline
\raisebox{-1.50ex}[0cm][0cm]{Depth }&
Resolution&
320$\times $240&
512$\times $424 \\
\cline{2-4}
 &
fps&
30fps&
30fps \\
\hline
\multicolumn{2}{|c|}{Player } &
6 &
6  \\
\hline
\multicolumn{2}{|c|}{Skeleton } &
2 &
6  \\
\hline
\multicolumn{2}{|c|}{Joint } &
20 Joint/ person&
25 Joint/person \\
\hline
\multicolumn{2}{|c|}{Range of Detection } &
0.8$\sim $4.0 m&
0.5$\sim $4.5 m \\
\hline
\raisebox{-1.50ex}[0cm][0cm]{Angle }&
Horizontal&
57 degree&
70 degree \\
\cline{2-4}
 &
Vertical&
43 degree&
60 degree \\
\hline
\multicolumn{2}{|c|}{Active IR video stream} &
NO&
512$\times $424,11-bit  \\
\hline
\end{tabular}
\label{tab1}
\end{table}

Dataset 1 (DUT RGB-D face dataset): This dataset utilizes Microsoft Kinect
for Xbox one V2.0 camera to acquire images, which acquires RGB images as
well as depth images. This dataset contains 1620 RGB-D (RGB and depth)
photos recorded with 6480 files of 54 people. Each class includes 30 faces. Expressions of happiness,
anger, sorrow, scare and surprise are acquired from five different angles
(up, down, left, right and middle) for each person. Color photos are
recorded with 8 bits, and each color image is decomposed into three files
(R, G, and B). Depths photos are recorded using 16-bit data to guarantee the
depth of facial small changes are accurately recorded. All people in these
photos do not wear glasses to ensure the precision of expression
acquisition.

Dataset 2: The RGB-D Household Object Dataset contains 300 household
objects. The dataset was captured by a Kinect V1 camera. Each image pair are
RGB and depth images (RGB size: 640$\times $480 and depth image at 30 Hz).
The objects fall into 51 categories. The objects are obtained by RGB-D video
from different angles of each object. More details can be referred to [16].

\subsection{Experimental results and analysis}

In social robot tasks, recognition and grasp are both important to
applications. The experimental results of DUT RGB-D face dataset are listed
in table 2 and figure 4 as follows:

\begin{table}[htbp]
\centering
\caption{ The comparison of DUT RGB-D face dataset ({\%})}
\begin{tabular}
{ccccc}
\hline
Methods&
RGB&
Depth&
Joint Fusion&
DSIFT+FGF \\
\hline
Recognition rate&
82.30&
60.01&
82.03&
84.50 \\
\hline
\end{tabular}
\label{tab2}
\end{table}

We use dense SIFT method to extract feature of RGB-D face dataset, and utilize one Vs. rest SVM classifer to complete the face recognition. In face recognition, we extract 3 training faces in each class and the rest as the testing set. As can be seen in table 2, depth information is more effective than RGB
representation. The RGB recognition rate is 82.30{\%} which is 22.29{\%}
higher than the depth faces. This is because that face recognition is high
related to RGB. An important result is that RGB+depth feature deduced
0.27{\%} than single RGB feature. This phenomenon illustrates joint feature
may suffer from data distribution changed. Our method achieve 84.50{\%}
which is higher than any single feature (or joint feature) and can deal with
the joint shortcoming by fused graph feature extraction.

\begin{center}
\begin{figure}[htbp]
\centerline{\includegraphics[width=3.5 in]{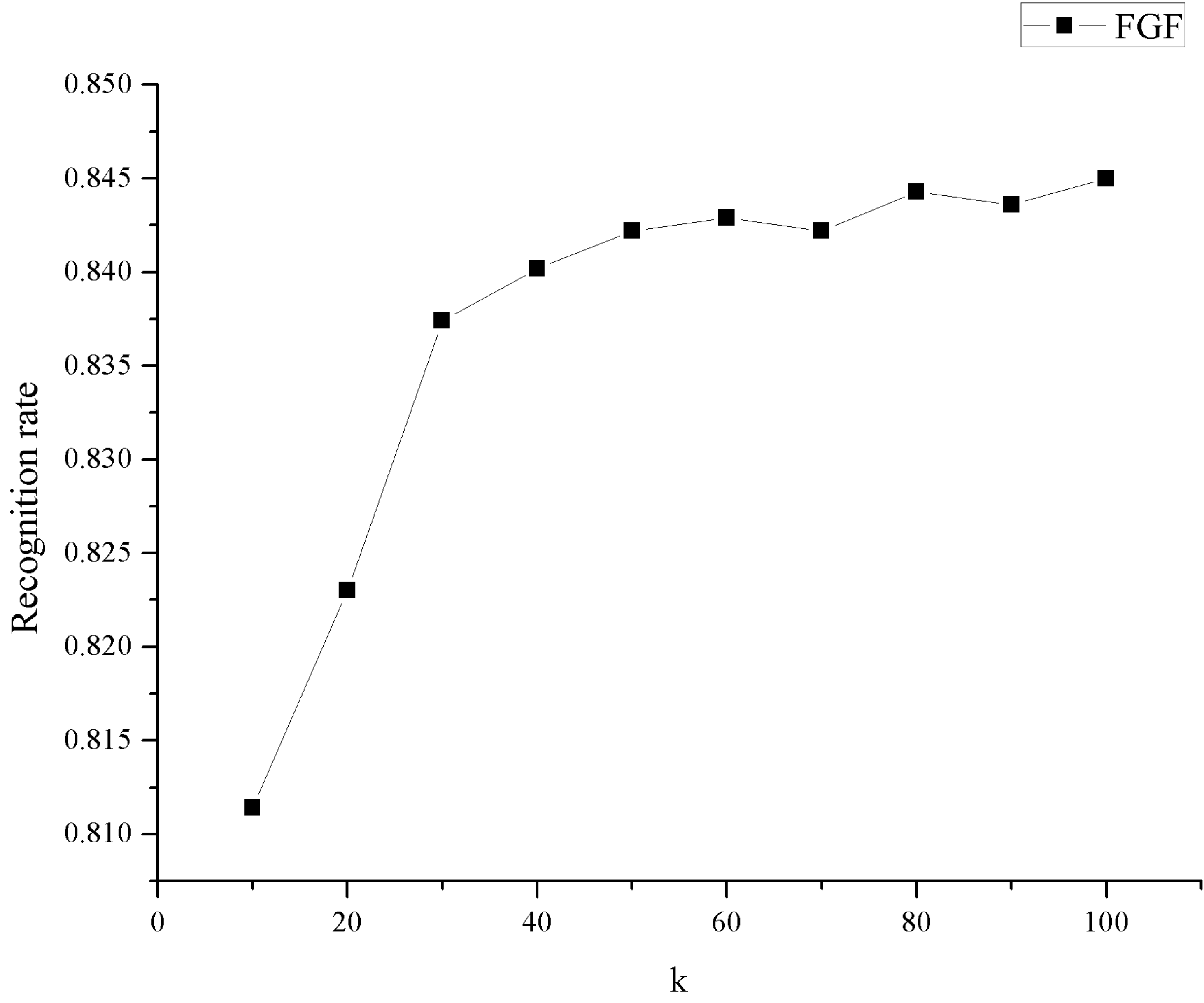}}
\caption{Recognition rate of FGF by changed k in DUT RGB-D face dataset}
\label{fig4}
\end{figure}
\end{center}

Fig. 4 shows parameter influence of our fusion model. We can see that FGF is
not sensitive to the change of parameter.


\begin{table*}[htbp]
\centering
\caption{The recognition rate of different splits in Dataset 2(\%)}
\begin{tabular}
{cccccccccccccc}
\hline
\raisebox{-1.50ex}[0cm][0cm]{Methods ($k$)}&
\raisebox{-1.50ex}[0cm][0cm]{$d$}&
\multicolumn{11}{c}{Recognition Rate} &
\raisebox{-1.50ex}[0cm][0cm]{std} \\
\cline{3-13}
 &
 &
s-1&
s-2&
s-3&
s-4&
s-5&
s-6&
s-7&
s-8&
s-9&
s-10&
mean&
  \\
\hline
crnn rgb&
8192&
92.47&
87.62&
88.73&
90.11&
89.57&
87.39&
90.85&
93.93&
89.06&
93.23&
90.30&
2.28 \\

crnn depth&
8192&
93.72&
94.95&
93.34&
92.24&
87.81&
92.21&
92.98&
94.82&
91.62&
96.06&
92.98&
2.29 \\

crnn rgb+depth&
16384&
93.95&
94.38&
93.26&
92.01&
88.05&
92.34&
93.53&
95.04&
91.68&
96.06&
93.03&
2.22 \\

crnn+FGF(50)&
50&
89.73&
92.28&
92.24&
89.35&
88.01&
92.27&
92.12&
91.39&
90.66&
95.25&
91.33&
2.02 \\

crnn+FGF(50)&
100&
89.82&
92.91&
92.53&
89.52&
88.27&
92.50&
93.45&
91.63&
90.67&
95.19&
91.65&
2.09 \\

crnn+FGF(50)&
200&
89.35&
94.41&
93.06&
91.13&
88.71&
92.52&
94.73&
91.85&
91.39&
96.02&
92.26&
2.22 \\

crnn+FGF(100)&
50&
91.59&
94.80&
93.56&
90.68&
88.62&
92.13&
95.81&
94.12&
92.24&
96.06&
92.96&
2.36 \\

crnn+FGF(100)&
100&
89.96&
95.26&
92.83&
91.40&
90.51&
93.22&
95.80&
92.66&
93.39&
96.21&
93.12&
2.15 \\

crnn+FGF(100)&
200&
91.65&
95.86&
93.69&
91.91&
89.57&
92.61&
96.03&
92.86&
93.29&
96.71&
93.42&
2.23 \\

crnn+FGF(150)&
50&
92.51&
95.21&
92.96&
95.14&
89.63&
92.35&
93.33&
92.89&
95.30&
98.00&
93.73&
2.27 \\

crnn+FGF(150)&
100&
92.09&
96.22&
94.00&
95.13&
90.33&
92.45&
92.32&
93.22&
94.56&
97.59&
93.79&
2.16 \\

crnn+FGF(150)&
200&
92.74&
96.03&
94.72&
95.34&
89.75&
92.34&
92.19&
93.79&
94.31&
98.02&
93.92&
2.32 \\
\hline
\end{tabular}
\label{tab3}
\end{table*}

\begin{center}
\begin{figure}[htbp]
\centerline{\includegraphics[width=3.5 in]{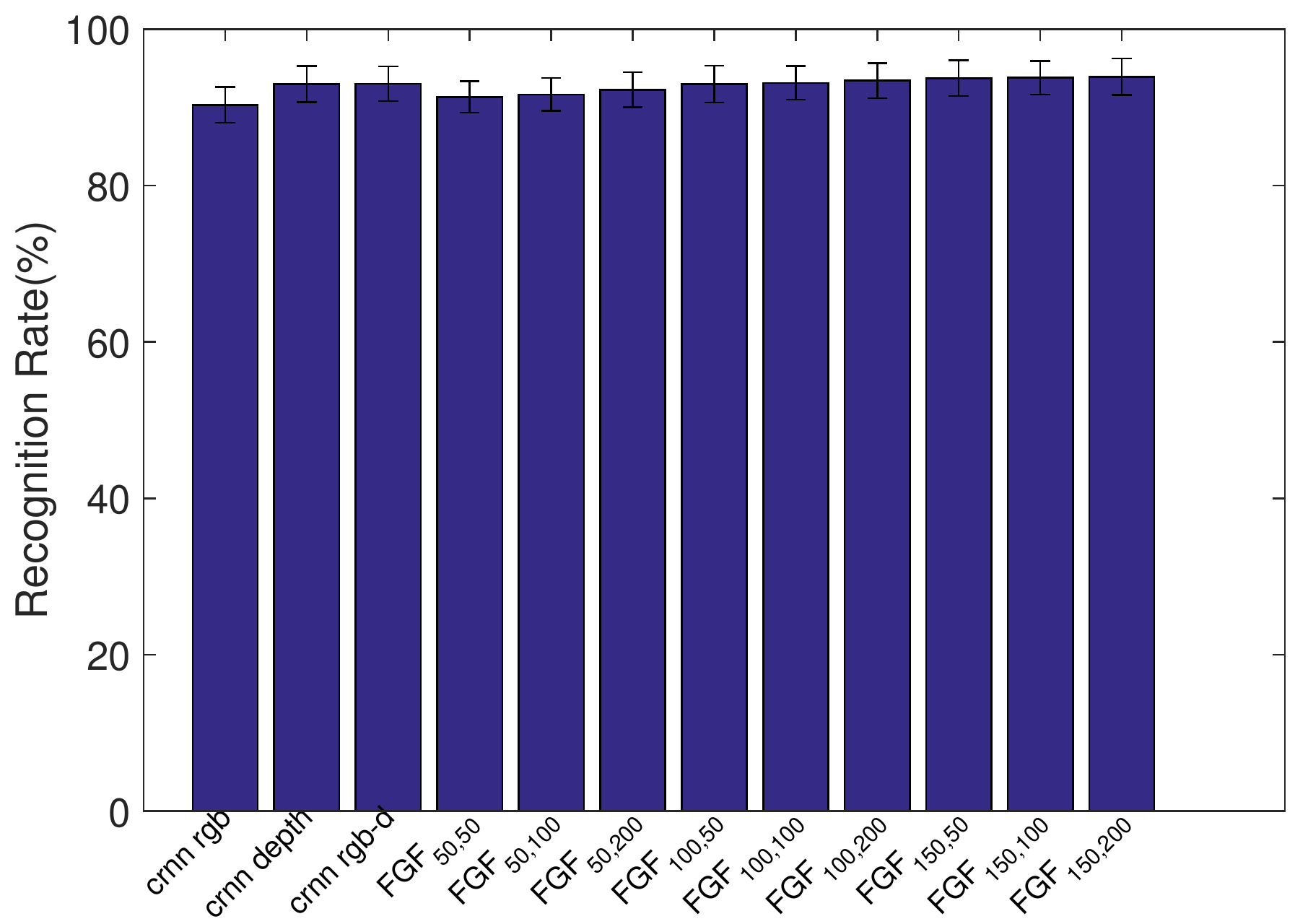}}
\caption{Recognition rate of CRNN and FGFs by changed $k$, $d$ in DUT RGB-D dataset2}
\label{fig5}
\end{figure}
\end{center}

In object experiment, we use CNN-RNN features extracting from RGB and depth images and split 10 times. Each split of testing set selects all images of one instance and the rest as training set. Table 3 shows the results of object RGB-D recognition. Different from face
experiments, object recognition using depth information can get 92.98{\%}
higher precision than that using RGB images. This result illustrates object
recognition more relies on ``depth feeling''. Our fused feature is more
effective and efficiency than the joint one (reach 93.92{\%} only using 200
dimension feature), though the joint feature enhances the precision in
object dataset. Table 4 and Fig. \ref{fig5}\footnote{$FGF_{k,d}$ denotes using different $k$ and $d$ in Dataset2.} illustrate that our method can achieve more higher results than other state-of-the-art methods.

\section{Conclusion}

In this paper, we built a vision robot with RGB-D camera and gave a DUT
RGB-D face dataset. We mainly proposed a RGB-D recognition method FGF and
evaluated FGF in two RGB-D datasets. FGF can get better performance than
previous approach and can help robot to execute complex tasks, such as SLAM,
compliant~hand designing, human-robot interaction etc.. We will consider
designing a more effective sensor and robust supervised dimensionality reduction method (such as reference \cite{29}) as robot vision in our future work.

\end{document}